\documentclass[letterpaper]{article} \usepackage[draft]{aaai25}  \usepackage{times}  \usepackage{helvet}  \usepackage{courier}  \usepackage[hyphens]{url}  \usepackage{graphicx}    \usepackage{natbib}  \usepackage{caption}    

\usepackage[utf8]{inputenc} \usepackage[T1]{fontenc}    \usepackage{url}            \usepackage{booktabs}       \usepackage{amsfonts}       \usepackage{nicefrac}       \usepackage{microtype}      \usepackage{xcolor}         \usepackage{mathtools}
\newcommand{\defeq}{\vcentcolon=}

\usepackage{algorithm}
\usepackage{algorithmic}
\usepackage{amssymb}
\usepackage{adjustbox}
\usepackage{graphicx}
\usepackage{amsthm}
\usepackage{amsmath}
\usepackage{cancel} 

\theoremstyle{plain}
\newtheorem{theorem}{Theorem}\newtheorem{claim}{Claim}\newtheorem{assumption}{Assumption}
\theoremstyle{definition}
\newtheorem{definition}{Definition}

\theoremstyle{remark}
\title{Boosting Test Performance with Importance Sampling--a Subpopulation Perspective}
\author{
Hongyu Shen\textsuperscript{\rm 1}, Zhizhen Zhao\textsuperscript{\rm 1}
}
\affiliations{
\textsuperscript{\rm 1} Department of Electrical and Computer Engineering,\\\
    University of Illinois at Urbana-Champaign, Champaign, IL, 61820, U.S.A. \\
    \{hongyu2, zhizhenz\}@illinois.edu

}

\usepackage{bibentry}

\begin{document}

\onecolumn
\maketitle

\begin{abstract}
Despite empirical risk minimization (ERM) is widely applied in the machine learning community, its performance is limited on data with spurious correlation or subpopulation that is introduced by hidden attributes. Existing literature proposed techniques to maximize group-balanced or worst-group accuracy when such correlation presents, yet, at the cost of lower average accuracy. In addition, many existing works conduct surveys on different subpopulation methods without revealing the inherent connection between these methods, which could hinder the technology advancement in this area. In this paper, we identify important sampling as a simple yet powerful tool for solving the subpopulation problem. On the theory side, we provide a new systematic formulation of the subpopulation problem and explicitly identify the assumptions that are not clearly stated in the existing works. This helps to uncover the cause of the dropped average accuracy. We provide the first theoretical discussion on the connections of existing methods, revealing the core components that make them different. On the application side, we demonstrate a single estimator is enough to solve the subpopulation problem. In particular, we introduce the estimator in both attribute-known and -unknown scenarios in the subpopulation setup, offering flexibility in practical use cases. And empirically, we achieve state-of-the-art performance on commonly used benchmark datasets.
\end{abstract}

\section{Introduction}\label{sec_intro}
\begin{figure}[htp]
  \centering
\includegraphics[width=0.35\textwidth]{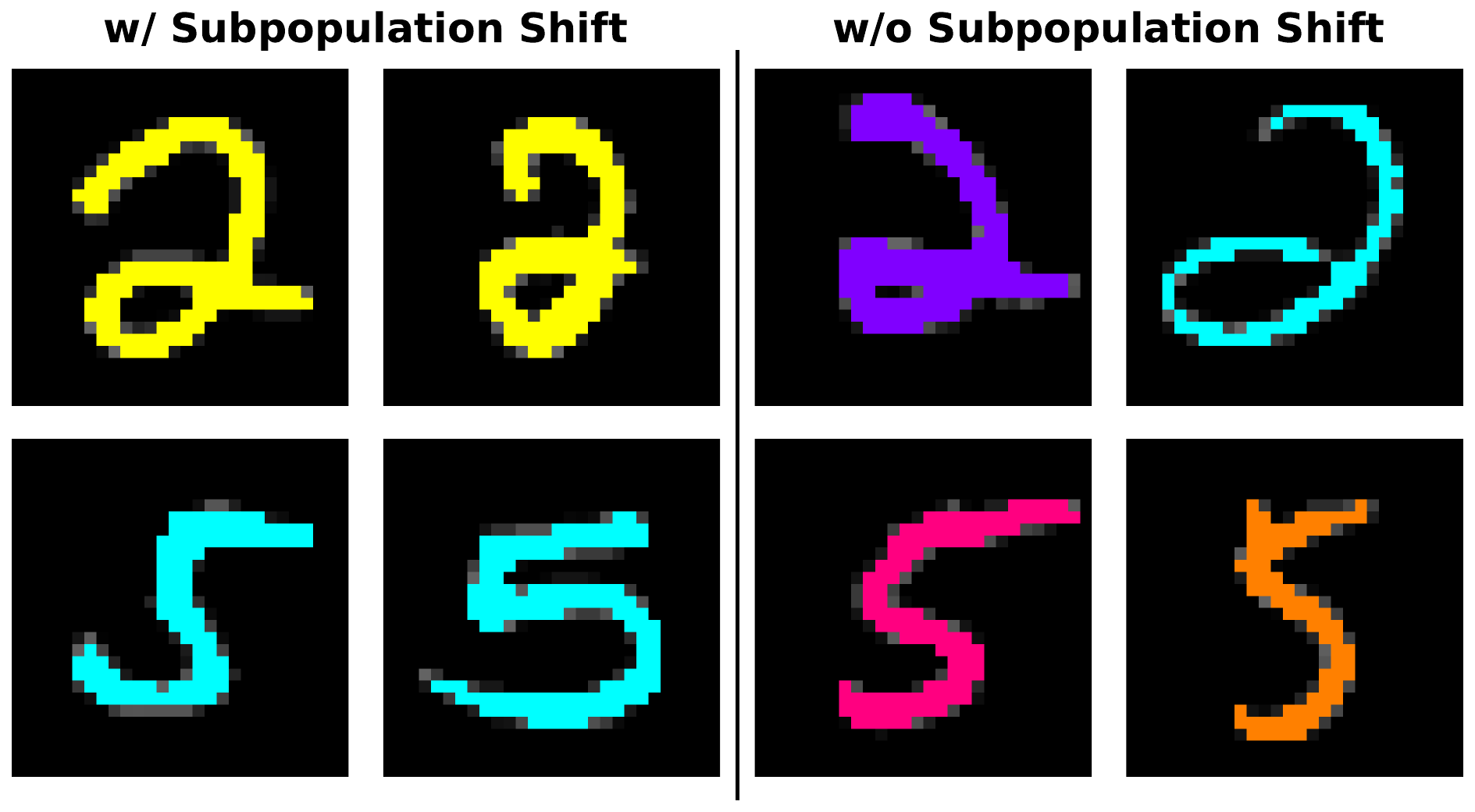} \caption{An image example on subpopulation shift. The left panel contains images where digits and colors are correlated, whereas the right panel does not exhibit such correlation.}
  \label{fig_example}
\end{figure}

Empirical risk minimization (ERM) often struggles with distribution shifts that manifest when the training and test distributions differ~\citep{bickel2007discriminative,DSIML,shimodaira2000improving}. One ubiquitous type of distribution shift is \emph{subpopulation
shift}, which describes a scenario where the portion of the subpopulations may vary between training and testing sets. See Figure~\ref{fig_example} for an example. This consequently leads to degraded performance when a trained model is applied to production/testing environments~\citep{yang2023change}. Ensuring that machine learning models are robust against these distribution shifts hence is crucial for their reliability and safe real-world application.

Existing works proposed different methods in the forms of auxiliary losses~\citep{li2018domain,arjovsky2019invariant,alshammari2022long}, data augmentations~\citep{zhang2017mixup,yao2022improving,han2022umix}, modeling objectives~\citep{liu2021just,sagawa2019distributionally,japkowicz2000class,wu2023discover,nam2020learning,asgari2022masktune,rudner2024mind,han2024improving,hong2023harmonizing,tsirigotis2024group,menon2020long,lin2017focal} and data sampling techniques~\citep{labonte2024towards,izmailov2022feature,japkowicz2000class}. They all exhibit superior performance on worst group accuracy while maintaining high accuracy in the overall set. However, two recent works experimentally observed that most models experience a drop in average accuracy performance compared to the ERM setup despite the high worst group accuracy~\citep{tsirigotis2024group,yang2023change}. Nonetheless, none of the papers is able to provide rigorous explanations on the answer to why. The lack of clarity in understanding can impede the development of appropriate models and methods, potentially stalling progress in the field.

In this work, we propose a systematic dataset bias analysis (DBA) framework that is rooted in importance sampling. With this framework, we reveal the cause of the lower-than-ERM average accuracy is the mismatch between the learning objective and the testing dataset. Moreover, we identify the flexibility of this framework in interpreting the formulation of some of the existing works that focus primarily on statistical heuristics and do not clearly specify the underlying assumptions of the models or data. The DBA framework, on the other hand, can close the gap, allowing us to explicitly discuss assumptions systematically and compare different existing works with the same language. We believe this analysis offers a comprehensive and theoretically grounded view to people who wish to proceed with the study of subpopulation methods.

Practically, we propose to estimate a single distribution given the conducted analysis using the DBA framework and prove that this is enough for solving the subpopulation problem under certain assumptions. Subsequently, we propose 3 different methods for estimating the distribution given different access levels to data and attributes. Empirically, we demonstrate the framework improves the test performance under subpopulation setups and achieves state-of-the-art (SOTA) results for both average and worst group accuracy while avoiding the the lower-to-ERM performance.

\section{Related Work}\label{sec_related_work_main}
In this section, we cover related works about importance sampling, the survey papers of the subpopulation shift, and the associated SOTA methods. Due to space limits, we only provide a concise version here and defer the complete version in Appendix~\ref{sec_related_work}.

\subsection{Importance Sampling}
DBA interprets distributional shift as a mismatch of the weight function from an importance sampling perspective. Although primarily focused on subpopulation setups, the method's formulation applies broadly to distributional shift problems. Early works on importance sampling~\cite{shimodaira2000improving, huang2006correcting} address dataset shifts but lack real-world experiments and clarity for subpopulation cases. In contrast, DBA systematically formulates the application to subpopulation problems, explicitly stating assumptions and identifying key components like distributions leading to such issues. Other studies~\cite{kanamori2009least, fang2020rethinking} propose weight estimation methods requiring partial test set access, unlike DBA. Additionally, DBA considers the weight function as the ratio of joint distributions of $x$ and $y$, addressing subpopulation and covariate shifts more realistically.

\subsection{Subpopulation Survey}
\citet{yang2023change} provides the first comprehensive experimental study on subpopulation methods. It uses Bayes' theorem to decompose $y\vert x$, accounting for attributes (spurious features), and categorizes datasets into four classes with varying label-attribute correlations. The paper benchmarks 20 subpopulation methods across these datasets but lacks statistical quantification of performance differences. Other surveys~\cite{yu2024survey,2312.16243} cover broader out-of-distribution (OOD) and domain generalization (DG) methods. While~\citet{yu2024survey} focuses on applications,~\citet{2312.16243} quantifies error inflation due to distribution shifts but doesn't address correction via model design. Our work extends prior studies by providing formal statistical analysis to quantify errors from both data and modeling perspectives. DBA also explains why some methods trade worst-case accuracy for lower average test accuracy.

\subsection{Subpopulation Method}
We categorize subpopulation methods into four classes: auxiliary losses, data augmentations, modeling objectives, and data sampling techniques. Auxiliary loss methods~\cite{li2018domain, arjovsky2019invariant, alshammari2022long} aim to mitigate the impact of spurious backgrounds via adversarial training, gradient regularization, or class-balanced adjustments. Data augmentation methods~\cite{zhang2017mixup, han2022umix, yao2022improving} use convex combinations of samples to reduce background effects. Data sampling methods~\cite{labonte2024towards, izmailov2022feature} identify class-balanced subsets with independent spurious features for finetuning. Modeling objective methods~\cite{sagawa2019distributionally, wu2023discover, lin2017focal, rudner2024mind} focus on robust feature learning, subpopulation correction, or tailored loss terms like KL divergence or mutual information. 

DBA stands out by explicitly stating data assumptions and connecting existing methods under a unified statistical framework (see Sec.~\ref{sec_connection}). For instance, it highlights that augmentation methods~\cite{zhang2017mixup, yao2022improving, han2022umix} assume conditional similarity across subpopulations. DBA also identifies a universal assumption of identical conditional generative models across methods, which previous works did not explicitly address. Empirically, DBA outperforms SOTA methods on three datasets, confirming its effectiveness and simplicity, and leveraging importance sampling for practical implementation.
 
\section{Method}
\label{sec_method}
The method section consists of two components: 1. the DBA framework; 2. three estimation methods for the only conditional outlined in the DBA framework, concerning three different scenarios where data assumptions vary. We first describe the framework, followed by the introduction of the proposed methods.

\subsection{Dataset Bias Analysis Framework}\label{sec_dba}

Throughout the paper, we consider the following notations: $x\in \mathcal{X}$ and $y\in \mathcal{Y}$ indicate the random variables for the data and labels, respectively. $\mathcal{X}$ and $\mathcal{Y}$ refer to their corresponding spaces. We denote $y$ as a discrete random variable. We use $p(\cdot)$ to denote the probability distribution and $q(\cdot)$ or $\hat{p}(\cdot)$ to represent the estimates. Subscripts ``tr'', ``va'', and ``te'' indicate concepts associated with train, validation, and test datasets, respectively. We use $\mathcal{D}$ to refer to the datasets. We let $\mathcal{M}_{\text{tr}} \defeq \{q(\cdot) \vert q(\cdot)\ \text{estimated with data in}\ \mathcal{D}_\text{tr} \}$ denote the model spaces for the general learning problem. $s$ denotes the attributes/spurious variables that are present in the datasets. This is also the root of the subpopulation. And $I$ refers to the dataset indicator, which is the abstract variable that has no real values (i.e. $I_\text{tr}$, $I_\text{va}$, and $I_\text{te}$). We use $\mathrm{Supp}(\cdot)$ to indicate the support set. We also use the notation ``$\sim$'' on two datasets (e.g. $\mathcal{D}_\text{tr} \sim \mathcal{D}_\text{te}$) to represent the same data distributions for the given datasets.

The DBA framework is formulated by initially asking the question: \emph{Which model do we pick after training?} Conventional approaches consider ERM over $\mathcal{D}_\text{tr}$, stop the training, and choose the model with the lowest loss value on $\mathcal{D}_\text{va}$. Usually, the losses are implicitly assumed to be identical across $\mathcal{D}_\text{tr}$, $\mathcal{D}_\text{va}$, and $\mathcal{D}_\text{te}$. There are two drawbacks to this inattentive assumption. First, it does not properly characterize the difference across different datasets. Second, it does not naturally take into account how people make choices on the model. As a remedy, we propose the following objective (Eq.~\eqref{eq_obj}) as the foundation for the DBA framework:
\begin{align}\label{eq_obj}
\mathbb{E}_{(x, y) \sim p(x, y\vert I_\text{va})} [\log q(y\vert x, I_\text{tr})].
\end{align}
The maximization of the objective (Eq.~\eqref{eq_obj_opt}) hence provides an intuitive view of how people choose the final model after the optimization:
\begin{align}\label{eq_obj_opt}
&\max_{q \in \mathcal{M}_\text{tr}} 
 \mathbb{E}_{(x, y) \sim p(x, y\vert I_\text{va})} [\log q(y\vert x, I_\text{tr})].
\end{align}
In this paper, we consistently focus on the predictive modeling setup (i.e. $y\vert x$), which is aligned with existing works. Intuitively, Eq.~\eqref{eq_obj_opt} describes the scenario where we find the best conditional predictive model $q$ according to the highest log likelihood measured over $\mathcal{D}_\text{va}$. Eq.~\eqref{eq_obj_opt} differs from ERM by explicitly considering the inherent difference between different datasets. In most cases, we seek for models to perform well on the unseen $\mathcal{D}_\text{te}$. To characterize this, we apply a similar logic as in Eq.~\eqref{eq_obj} and focus on measuring the difference between validation and test sets. We make the following universal assumption~\ref{assume_support}.
\begin{assumption}\label{assume_support}
The supports of $x$, $y$ on $\mathcal{D}_\text{tr}$, $\mathcal{D}_\text{va}$, and $\mathcal{D}_\text{te}$ follow the relationship: \\
$\mathrm{Supp}_\text{tr}(x, y) \supset \mathrm{Supp}_\text{va}(x, y)$,  $\mathrm{Supp}_\text{tr}(x, y) \supset \mathrm{Supp}_\text{te}(x, y)$, and $\mathrm{Supp}_\text{va}(x, y) \supset \mathrm{Supp}_\text{te}(x, y)$.
\end{assumption}
The inclusion relationship described in the Assumption~\ref{assume_support} essentially ensures a well-defined weight function (i.e., the denominator of the weight function is not zero) in the importance sampling setup in the proposed DBA framework.
With this assumption, we make the following claim on the performance of the picked model (from Eq.~\eqref{eq_obj_opt}) with $\mathcal{D}_\text{te}$: \emph{How does the picked model perform on the test set?}
\begin{claim}\label{thm_test_val}
Given Assumption~\ref{assume_support} holds and let $q^*$ denote the best model obtained from Eq.~\eqref{eq_obj_opt}. The likelihood evaluated with the test set $\mathcal{D}_\text{te}$ for the model $q^*$ can be viewed as the importance sampling version of $\mathbb{E}_{(x, y) \sim p(x, y\vert I_\text{va})} [z(x, y, I_\text{va}, I_\text{te})\log q^*(y\vert x, I_\text{tr})]$ over the validation set with the function $z(\cdot)$ defined below:
\begin{align}\label{eq_val_test_proposal}
z(x, y, I_\text{va}, I_\text{te}) \defeq \frac{p(x, y\vert I_\text{te})}{p(x, y\vert I_\text{va})}.
\end{align}
\end{claim}
We defer this and all the following proof details in Appendix~\ref{ap_proof}. Claim~\ref{thm_test_val} informs that the only way to guarantee the best testing performance for the picked model $q^*$ is to have access to the distribution $p(x, y\vert I_\text{te})$. This points out a hidden pitfall that commonly exists, yet overlooked, in the current machine learning optimizations with ERM---people choose a model with the best validation performance and report the corresponding testing performance. By Claim~\ref{thm_test_val}, we know that this general setup is true only in the case where $p(x, y\vert I_\text{va})=p(x, y\vert I_\text{te})$. Otherwise, one needs to provide an accurate estimation on $z(x, y, I_\text{va}, I_\text{te})$ and pick the training model via a weighted likelihood, $\mathbb{E}_{(x, y) \sim p(x, y\vert I_\text{va})} [z(x, y, I_\text{va}, I_\text{te})\log q^*(y\vert x, I_\text{tr})]$, on the validation set, to achieve optimal performance on the test set.

Simply put, Eq.~\eqref{eq_obj_opt} describes the way people pick the model during optimization, and Claim~\ref{thm_test_val} points out the correct picking criterion for maximum test set performance. A natural follow-up question on these two arguments is: \emph{Can we combine the notion of training and picking, and directly optimize $q$ to maximize the testing performance?} The answer is affirmative under some additional assumptions. To explain, we first claim an optimization equivalence, providing the general form with which the optimization on the training set is identical to the optimization on the testing set (Claim~\ref{thm_adjust_obj_general}). Then we derive another objective in the setup where we obtain a closed-form $g(x, y, I_\text{tr}, I_\text{te})$ (see Claim~\ref{thm_adjust_obj_general}) after making several assumptions on the structure of the test data (Theorem~\ref{thm_adjust_obj_assumption}).

\begin{claim}\label{thm_adjust_obj_general}
Given Assumption~\ref{assume_support} holds we obtain the following equality on the objective:
\begin{align}\label{eq_adjust_obj}
& \mathbb{E}_{(x, y) \sim p(x, y\vert I_\text{te})} [\log q(y\vert x, I_\text{tr})] \nonumber \\
 = & \mathbb{E}_{(x, y) \sim p(x, y\vert I_\text{tr})} [g(x, y, I_\text{tr}, I_\text{te})\log q(y\vert x, I_\text{tr})],
\end{align}
where the weight function $g(x, y, I_\text{tr}, I_\text{te}) \defeq \frac{p(x, y\vert I_\text{te})}{p(x, y\vert I_\text{tr})}$.
\end{claim}
The proof is similar to Claim~\ref{thm_test_val} and can be found in Appendix~\ref{ap_proof}. Note that in the language of importance sampling, the weight function $g(x, y, I_\text{tr}, I_\text{te})$ consists of the proposal distribution $p(x, y\vert I_\text{tr})$ and the data distribution $p(x, y\vert I_\text{te})$ in our setup. Compared to Eq.~\eqref{eq_obj}, Eq.~\eqref{eq_adjust_obj} offers an objective that can be optimized with $\mathcal{D}_\text{tr}$ as the expectation is taken over the training set---the same space defined for models $q \in \mathcal{M}_\text{tr}$. Claim~\ref{thm_adjust_obj_general} also confirms that one must know $p(x, y\vert I_\text{te})$ to improve the testing performance of $q$ when optimizing a model.

In this paper, we consider a uniform attribute setup that assumes the uniform distribution on the attribute/spurious variable $s \in \mathcal{S}$, which is a discrete random variable and $\mathrm{Supp}(s)=\mathrm{Supp}(y)$. $s$ represents the cause of the subpopulation in our study. Formally speaking, this paper considers the following subpopulation shift:
\begin{definition}\label{definition_subpopulation}
The subpopulation shift is defined as the distributional difference between $p(x, y\vert I_\text{tr})$ and $p(x, y\vert I_\text{te})$ that is introduced by the spurious variable $s$ w.r.t. the response $y$. Namely, $p(s, y\vert I_\text{tr}) \neq p(s, y\vert I_\text{te})$. \end{definition}
Specifically, we decompose the joint distribution of $x$ and $y$ through $\sum_s p(x, y, s\vert I_\text{tr}) =\sum_s p(x\vert y, s, I_\text{tr})p(y, s\vert I_\text{tr})$, and $\sum_s p(x, y, s\vert I_\text{te}) =\sum_s p(x\vert y, s, I_\text{te})p(y, s\vert I_\text{te})$. And the difference between datasets is on $p(s, y\vert I_\text{tr}) \neq p(s, y\vert I_\text{te})$.
In the following, we describe several assumptions that lead to the major result of the paper---Theorem~\ref{thm_adjust_obj_assumption}:
\begin{assumption}\label{assume_data_gen}
A universal data generator given the dataset information $I$, the label $y$, and the attribute $s$ for the training and test sets: $p(x\vert y, s, I_\text{tr}) = p(x\vert y, s, I_\text{te})$.
\end{assumption}
\begin{assumption}\label{assume_s_uniform}
The attribute variable $s$ follows a uniform distribution, conditional on $y$ and $ I_\text{te}$: $p(s\vert y, I_\text{te}) = 1/L$, where $L$ is the number of outcomes for the discrete random variable $s$.
\end{assumption}
Assumption~\ref{assume_data_gen} requires identical generative processes for \(x\) across training and testing. This can be seen as a specific type of covariate shift, attributing shifts in \(p(x, y)\) to variations in \(p(y, s)\) given the attribute $s$, rather than \(p(x)\). Such an assumption is common in conformal analysis and causal inference~\citep{yang2024doubly,suter2019robustly,lei2021conformal}. 
Assumption~\ref{assume_s_uniform} imposes a weaker assumption compared to the literature, where uniformity and independence are generally assumed for both $y$ and $s$~\citep{tsirigotis2024group}. Compared to the existing work, we only assume $s$ to follow a uniform distribution and there is no constraint on the distribution of $y$. The latter makes this approach applicable to class-imbalanced test data. 

We further make two additional assumptions (Assumption~\ref{assume_proposal_1} and~\ref{assume_proposal_2}) that reflect the nature of the considered subpopulation problems. This starts with studying the composition of the shifted datasets. Specifically, we introduce a random variable $m$ that explicitly describes the substructure of the given data (i.e. $\mathcal{D}_\text{tr}$, $\mathcal{D}_\text{va}$, and $\mathcal{D}_\text{te}$). Most existing works only consider the attribute variable $s$ and its relation to labels $y$ and data $x$. However, we realize that simply introducing this attribute is not enough to quantify the subpopulation as different subpopulations may have distinct relationships between $s$, $y$, and $x$. Therefore, the presence of $m$ enables the quantification of such differences, making the proposed framework more flexible. 

In particular, we consider $m$ to be a binary random variable that takes values $m_0$ or $m_1$. And $m_0$ refers to the conceptual minority group in $\mathcal{D}_\text{tr}$ that shares the same statistics for $s$, $y$, and $x$ in $\mathcal{D}_\text{te}$, whereas $m_1$ denotes the majority group that has distinct statistics of $s$, $y$, and possibly $x$---this explicitly characterizes the prevalent subpopulation in $\mathcal{D}_\text{tr}$ that causes the underperformance in $\mathcal{D}_\text{te}$. One may question the soundness of why we claim it is possible to find such a minority group in $\mathcal{D}_\text{tr}$. An intuitive, yet not strict, proof is to consider the established Assumption~\ref{assume_support} that constrains inclusive supports across datasets. With Assumption~\ref{assume_support}, we can always find a subset of $\mathcal{D}_\text{tr}$ whose data statistics are close to that of $\mathcal{D}_\text{te}$ for any possibly large enough datasets. This leads to the following assumption:
\begin{assumption}\label{assume_proposal_1}
$p(y\vert I_\text{te}) = p(y\vert m_0, I_\text{tr}) = p(y\vert I_\text{tr})$.
\end{assumption}
Assumption~\ref{assume_proposal_1} describes the scenario where there is no subpupolation on $y$ between $\mathcal{D}_\text{tr}$ and $\mathcal{D}_\text{te}$. This assumption indicates that the subpopulation is introduced by the association between $s$ and $y$, or $x$ and $y$, but not solely by $y$ itself. Since the minority group $m_0$ shares same data statistics as $y$, it is natural to have the equality $p(y\vert I_\text{te}) = p(y\vert m_0, I_\text{tr})$. It is noteworthy that there is no constraint on the number of groups specified by $m$. The size of 2 is considered in this paper due to its simplicity and high performance in practice (see Sec.~\ref{sec_exp}). 

Assumption~\ref{assume_proposal_2}, on the other hand, quantifies explicitly that there is a portion (i.e. $m_1$) of samples in $\mathcal{D}_\text{tr}$ whose attributes $s$ are identical to the labels $y$. Rather than treating it as an assumption, it is more of a characterization on the subpopulation that widely presents in the real-world data (e.g., Waterbirds and ColorMNIST, or others described in~\cite{yang2023change}), where attributes strongly mislead the model prediction by such correlation.
\begin{assumption}\label{assume_proposal_2}
$p(s\vert y, m_1, I_\text{tr}) = \mathbf{1}_{\{y=s\}}$, where $\mathbf{1}_{\{y=s\}}$ is the indicator function.
\end{assumption}

With all ingredients, we propose the following theorem on the modeling objective:
\begin{theorem}\label{thm_adjust_obj_assumption}
Given Assumption~\ref{assume_support},~\ref{assume_data_gen},~\ref{assume_s_uniform},~\ref{assume_proposal_1}, and~\ref{assume_proposal_2} hold, the optimization of Eq.~\eqref{eq_adjust_obj} with the following weight function $g(x, y, I_\text{tr}, I_\text{te})$ directly maximizes the testing performance:
\begin{align}\label{eq_adjust_obj_assumption}
& g(x, y, I_\text{tr}, I_\text{te})^{-1} \defeq  p(m_0\vert I_\text{tr}) \nonumber \\
+ &  \frac{p(m_1\vert I_\text{tr}) \cdot \frac{L}{p(y\vert I_\text{tr})} \cdot p(y\vert m_1, I_\text{tr})}{ 1 + \left[ \frac{p(m_0\vert I_\text{tr})\cdot p(y\vert I_\text{tr})/L + p(y\vert m_1, I_\text{tr})}{ p(m_0\vert I_\text{tr})\cdot p(y\vert I_\text{tr})/L} \right] \cdot \frac{1 - p(s=y\vert y, x, I_\text{tr})}{p(s=y\vert y, x, I_\text{tr})}},
\end{align}
where $p(y\vert m_1, I_\text{tr}) = \frac{p(y\vert I_\text{tr}) - p(m_0\vert I_\text{tr})\cdot p(y\vert I_\text{tr})}{p(m_1\vert I_\text{tr})}$. $p(m_0\vert I_\text{tr})$ and $p(m_1\vert I_\text{tr})$ represent the probability of a binary random variable $m$ taking the value $m_0$ or $m_1$, respectively. Namely, the random variable $m$ denotes the split of $\mathcal{D}_\text{tr}$ into the majority and minority groups. 
\end{theorem}
The corresponding proof can be found in Appendix~\ref{ap_proof}. 
With this formulation, $p(s=y\vert y, x, I_\text{tr})$ is the only unknown term to be estimated. Theorem~\ref{thm_adjust_obj_assumption} provides a closed form objective with which models trained with $\mathcal{D}_\text{tr}$ perform optimally on $\mathcal{D}_\text{te}$. In the following, we consider 3 different setups on the accessibility of $s$ and the relationship between $\mathcal{D}_\text{tr}$ and $\mathcal{D}_\text{va}$. In each setup, we provide a method to estimate Eq.~\eqref{eq_adjust_obj_assumption}. We further showcase the performance of the proposed methods in the experiment section (Sec.~\ref{sec_exp}). In Appendix~\ref{sec_relaxation}, we include a discussion on the limitations of this approach concerning the restriction and possible relaxation of the assumptions.

\subsection{Dataset Bias Correction Method}\label{sec_bcm}
In this section, we provide 3 different approaches to estimate $Eq.~\eqref{eq_adjust_obj_assumption}$. We summarize these approaches with a general name: dataset bias correction method (DBCM). The 3 approaches essentially provide different ways of estimating the only missing term $p(s\vert y, x, I_\text{tr})$ in Eq.~\eqref{eq_adjust_obj_assumption}. Once the term is estimated, we employ a universal algorithm (see Algorithm~\ref{alg_train}) to train the model with $\mathcal{D}_\text{tr}$. To facilitate the use of this approach in more real-world applications, we describe the scenarios where the three following approaches can be applied in Appendix~\ref{sec_app_diff_scenarios}.

\subsubsection{Attribute $s$ is Known}\label{sec_known_s}
When we have access to the attribute $s$, we can make a direct estimation on the only unknown term $p(s=y\vert y, x, I_\text{tr})$ using the data $(x, y, s) \in \mathcal{D}_\text{tr}$ and apply Algorithm~\ref{alg_train} therein. As \(p(s=y|y, x, I_{\text{tr}})\) increases, the weight function \(g\) decreases (see Eq.~\eqref{eq_adjust_obj_assumption}), because stronger spurious correlations make \(p(s=y|y, x, I_{\text{tr}})\) larger. Down-weighting these samples during training helps performance by reducing reliance on spurious correlations.

\subsubsection{Attribute $s$ is Unknown and $\mathcal{D}_\text{tr} \sim \mathcal{D}_\text{va}$} \label{sec_unknown_s_same_data}
When we do not have access to the attribute $s$ and  $\mathcal{D}_\text{tr} \sim \mathcal{D}_\text{va}$, we propose to use the following term to estimate $p(s\vert y, x, I_\text{tr})$:
\begin{align}\label{eq_s_unknow_same_data}
 \hat{p}(s=y\vert y, x, I_\text{tr}) \propto \exp \left(\frac{\vert \log \hat{p}(y\vert x, I_\text{tr}) - \log \hat{p}(y\vert x, I_\text{va})\vert}{\tau} \right)^{-1},
\end{align}
where $\hat{p}(y\vert x, I_\text{tr})$ and $\hat{p}(y\vert x, I_\text{va})$ are the predictive models learned with $\mathcal{D}_\text{tr}$ and $\mathcal{D}_\text{va}$, respectively. And $\tau$ is the temperature hyperparameter. In practice we find $\tau=1$ consistently produces good results. We explicitly introduce $\tau$ to allow flexibility in the control of the estimation in Eq.~\eqref{eq_s_unknow_same_data}. Specifically, we first overfit two independent predictive models on both $\mathcal{D}_\text{tr}$ and $\mathcal{D}_\text{va}$ and then measure the difference on the two approximate laws with the training data. Note that $p(s=y\vert y, x, I_\text{tr})$ captures how likely $s$ shares the same label as $y$, which is the only unknown term evaluated in Eq.~\eqref{eq_adjust_obj_assumption}. Therefore, we do not need to recover the full distribution $p(s\vert y, x, I_\text{tr})$. Instead, we only need to quantify $\hat{p}(s=y\vert y, x, I_\text{tr})$---``how likely the bias is biased towards the true label $y$.'' This is captured by Eq.~\eqref{eq_s_unknow_same_data}, as if two models (trained separately on training and validation data) produce similar likelihoods (i.e. the difference in Eq.~\eqref{eq_s_unknow_same_data} is smaller) on a given input, then the input must associate with the attribute $s$ that is same as $y$. To summarize this approach in one line: \emph{two overfitted models act as a bias corrector!}

\subsubsection{Attribute $s$ is Unknown and $\mathcal{D}_\text{tr} \nsim \mathcal{D}_\text{va}$} \label{sec_unknown_s_different_data}
On the other hand, when $\mathcal{D}_\text{tr} \nsim \mathcal{D}_\text{va}$, we cannot utilize the predictive model estimated with $\mathcal{D}_\text{va}$. Instead, we propose to use the following term as an alternative,
\begin{align}\label{eq_s_unknown_diff_data}
 \hat{p}(s=y\vert y, x, I_\text{tr}) \propto \exp\left(-\frac{\log \hat{p}(y\vert x, I_\text{tr})}{\tau}\right)^{-1}.
\end{align}
This is according to the observation that machine learning models tend to learn the correlated attributes $s$ with $y$ easily~\citep{asgari2022masktune}. In our case, we simply use $\hat{p}(y\vert x, I_\text{tr})$ as the proxy to characterize such correlation. In this case, samples with high accuracy should be down-weighted, as the model easily learns spurious correlations.

\subsection{Choose Models}\label{sec_stopping_criteria}

Similarly, we discuss different approaches for choosing a model. Unlike conventional methods that consistently use $\mathcal{D}_\text{va}$ to decide which model to choose, we propose to consider different ways for choosing a model when relationships between $\mathcal{D}_\text{va}$ and $\mathcal{D}_\text{te}$ are different. When $\mathcal{D}_\text{va} \sim \mathcal{D}_\text{te}$, according to Eq.~\eqref{eq_val_test_proposal}, $z(x, y, I_\text{va}, I_\text{te})=1$. This indicates that evaluating models on validation set is equivalent to evaluating on the test set, which corresponds to the conventional approach. However, things change when $\mathcal{D}_\text{va} \nsim \mathcal{D}_\text{te}$. This suggests that $\mathcal{D}_\text{va}$ is not sufficient in measuring the model performance for the test set as $z(x, y, I_\text{va}, I_\text{te}) \ne 1$. In this case, we can adopt the similar approach outlined in Sec.~\ref{sec_unknown_s_different_data} to estimate $z(x, y, I_\text{va}, I_\text{te})$, which focuses on $\mathcal{D}_\text{va}$ and $\mathcal{D}_\text{te}$, rather than $\mathcal{D}_\text{tr}$ and $\mathcal{D}_\text{te}$.

\begin{figure}
\begin{minipage}{1.\textwidth}
\begin{algorithm}[H] 
\textbf{Input} The initialized model $q(y\vert x, I_\text{tr})$; dataset $\mathcal{D}_\text{tr}$; The estimation $\hat{p}(s\vert y, x, I_\text{tr})$. \\
\textbf{Output:} the optimized $q(y\vert x, I_\text{tr})$.
\begin{algorithmic}[1]
\STATE Obtain $\hat{g}(x, y, I_\text{tr}, I_\text{te})$ given $\hat{p}(s=y\vert y, x, I_\text{tr})$ (see Eq.~\eqref{eq_adjust_obj_assumption}).
\STATE Perform the following optimization using $\mathcal{D}_\text{tr}$:
\begin{align}\label{eq_obj_general_algo}
{\max_{q \in \mathcal{M}_\text{tr}} \mathbb{E}_{(x, y) \sim p(x, y\vert I_\text{tr})} [\hat{g}(x, y, I_\text{tr}, I_\text{te}) \log q(y\vert x, I_\text{tr})].}
\end{align}
\end{algorithmic}
\caption{The universal algorithm for optimizing $q(y\vert x, I_\text{tr})$.}\label{alg_train}
\end{algorithm}
\end{minipage}
\end{figure}

\section{DBA Interpretation on Existing Work}\label{sec_connection}
In this section, we showcase how some representative existing works can be related to the DBA framework. Such discussion should complement the existing survey papers on subpopulation/distributional shifts and provide insights on the methodological development in the future. We follow the previously introduced categorization.

\texttt{The model objective class:}~\citet{liu2021just} and~\citet{nam2020learning} can be viewed as proposing different forms of the $p(s\vert y, x, I_\text{tr})$ estimation, where the former utilizes the classification accuracy and the latter considers generalized cross-entropy. Sec.~\ref{sec_unknown_s_same_data} and~\ref{sec_unknown_s_different_data} provide rationale on the validity of these terms---essentially they characterize the probability $p(s=y\vert y, x, I_\text{tr})$. Besides the variants of $\hat{p}(s=y\vert y, x, I_\text{tr})$, they propose different training schemes to correct.~\citet{liu2021just} subsamples the training set with their $\hat{p}(s=y\vert y, x, I_\text{tr})$ and~\citet{nam2020learning} proposes a parallel model to reweight samples according to $\hat{p}(s=y\vert y, x, I_\text{tr})$ from the generalized cross entropy. Nonetheless, none of them is alike DBCM, which is statistically consistent in directly improving the testing performance. 

\texttt{The data sampling class:} The methods in the data sampling class share great similarity to ours, as the proposed DBCM is essentially an importance sampling (reweighing) mechanism. ReWeight and ReSample~\citep{japkowicz2000class} can be treated as variants of the sampling technique. Precisely, ReWeight adjusts each sample weight according to the class ratio, in order to recover the class-balanced setup. Similarly, ReSample bootstraps the dataset with class-balanced weights. Essentially, they can be treated as the direct estimation of $g(x, y, I_\text{tr}, I_\text{te}) \defeq \frac{p(x, y\vert I_\text{te})}{p(x, y\vert I_\text{tr})}$, assuming $p(y\vert I_\text{tr})$ is uniform. When considering the presence of attribute $s$, $g(x, y, I_\text{tr}, I_\text{te})$ becomes,
\begin{align}\label{eq_sampling_class}
g(x, y, I_\text{tr}, I_\text{te}) & \defeq \frac{p(x, y\vert I_\text{te})}{p(x, y\vert I_\text{tr})} = \frac{\sum_s p(x\vert y, s, I_\text{te})p(y, s\vert I_\text{te})}{\sum_s p(x\vert y, s, I_\text{tr})p(y, s\vert I_\text{tr})}.
\end{align}
Their setups, in this case, further assume $p(y, s\vert I_\text{te})$ is uniform and $p(y, s\vert I_\text{tr})=p(y, s\vert I_\text{te})$, which is a stronger assumption compared to the proposed.

\texttt{The auxiliary loss class:}~\citet{tsirigotis2024group} and~\citet{menon2020long} are commonly used logit adjustment methods. With the DBA framework, they can be viewed as a two-step method. First, both methods propose an estimation of $p(y, s=y\vert, x, I_\text{tr})$. Then the estimates are used as a penalty term to regularize the ERM of the predictive model $q(y\vert x, I_\text{tr})$. In the first step,~\citet{tsirigotis2024group} applies a similar approach to one described in~\citep{liu2021just}. Both share conceptual similarity to the DBCM variant in Sec.~\ref{sec_unknown_s_different_data}.~\citet{menon2020long}, on the other hand, simply enforces the uniform  class balance assumption. Once $\hat{p}(y, s=y\vert x, I_\text{tr})$ is obtained, they optimize w.r.t.
\begin{align}\label{eq_aux_class}
\mathbb{E}_{(x, y) \sim p(x, y\vert I_\text{tr})} [\log q(y\vert x, I_\text{tr}) + \log\hat{p}(y, s=y\vert x, I_\text{tr})].
\end{align}
To compare the difference between Eq.~\eqref{eq_aux_class} and the optimal objective (Eq.~\eqref{eq_adjust_obj}), we prove the following theorem with two additional assumptions on the label $y$.
\begin{assumption}\label{assume_aux_y}
The label $y$ given $I_\text{tr}$ follows a uniform distribution.
\end{assumption}
\begin{assumption}\label{assume_aux_maj}
The training set contains only the dominant group $m_1$: $p(m_1\vert I_\text{tr}) = 1$.
\end{assumption}
\begin{theorem}\label{thm_aux_class}
Given Assumption~\ref{assume_support},~\ref{assume_data_gen},~\ref{assume_s_uniform}~\ref{assume_aux_y}, and~\ref{assume_aux_maj} hold, the optimization of Eq.~\eqref{eq_adjust_obj} with the following weight function $g(x, y, I_\text{tr}, I_\text{te})$ directly maximizes the testing performance:
\begin{align}\label{eq_aux_g}
g(x, y, I_\text{tr}, I_\text{te})^{-1} \defeq L\cdot p(y, s=y\vert x, I_\text{tr}).
\end{align}
And the objective Eq.~\eqref{eq_adjust_obj} is of form:
\begin{align}\label{eq_aux_obj}
& \mathbb{E}_{(x, y) \sim p(x, y\vert I_\text{tr})} \big[ g(x, y, I_\text{tr}, I_\text{te}) \big( \log q(y\vert x, I_\text{tr}) \nonumber \\ 
+ & \log p(y, s=y\vert x, I_\text{tr})\big) 
 + g(x, y, I_\text{tr}, I_\text{te}) \log L\cdot g(x, y, I_\text{tr}, I_\text{te}) \big].
\end{align}
\end{theorem}
The proof is deferred to Appendix~\ref{ap_proof}. Eq.~\eqref{eq_aux_class} differs Eq.~\eqref{eq_aux_obj} by 2 aspects. First, Eq.~\eqref{eq_aux_class} ignores the weight function $g(x, y, I_\text{tr}, I_\text{te})$ before the summation. Second, the regularization $g(x, y, I_\text{tr}, I_\text{te}) \log L\cdot g(x, y, I_\text{tr}, I_\text{te})$ in Eq.~\eqref{eq_aux_obj} is missing. Without these terms, Eq.~\eqref{eq_aux_obj} is not guaranteed to optimize for a class-balance dataset, as indicated in~\citep{tsirigotis2024group,menon2020long}. Consequently, these methods may underperform.

\texttt{The augmentation class:} Despite existing works provide augmentation techniques in the form of linear combination~\citep{zhang2017mixup,yao2022improving,han2022umix}, none of the papers provides statistical interpretation on why such techniques work better than ERM. We see our DBA framework as the first to provide supports for the soundness of the augmentation technique. In short, the augmentation to combine data samples can be viewed as variations of the direct recovery of $g(x, y, I_\text{tr}, I_\text{te}) \defeq \frac{p(x, y\vert I_\text{te})}{p(x, y\vert I_\text{tr})}$ under a different set of assumptions. Specifically, we provide the following Theorem~\ref{thm_aug_class} to support this statement. The proof can be found in Appendix~\ref{ap_proof}. In the following theorem, $m_0$ and $m_1$ are identical to the terms introduced in Theorem~\ref{thm_adjust_obj_assumption}. We first describe the assumptions.
\begin{assumption}\label{assume_x_dist_equal}
The data generator of $\mathcal{D}_\text{tr}$ are conditionally identical given different group information $m$: $p(x\vert m_0, I_\text{tr})=p(x\vert m_1, I_\text{tr})=p(x\vert I_\text{tr})$.
\end{assumption}
\begin{assumption}\label{assume_y_given_x_minor_match}
The predictive model on $\mathcal{D}_\text{te}$ shares the same law with the model that is conditioned on the group $m_0$ for $\mathcal{D}_\text{tr}$: $p(y\vert x, I_\text{te}) = p(y\vert x, m_0, I_\text{tr})$.
\end{assumption}
Note that Assumption~\ref{assume_y_given_x_minor_match} is conceptually similar to the setup for Theorem~\ref{thm_adjust_obj_assumption}.

\begin{theorem}\label{thm_aug_class}
Given Assumption~\ref{assume_support},~\ref{assume_x_dist_equal}, and~\ref{assume_y_given_x_minor_match} hold, the weight function $g(x, y, I_\text{tr}, I_\text{te})$ has the following form:
\begin{align}\label{eq_aug_g}
g(x, y, I_\text{tr}, I_\text{te})^{-1} \defeq & \lambda_0(x, I_\text{tr}, I_\text{te}) \cdot p(x\vert I_\text{tr}) \nonumber \\
+ & \lambda_1(x, y, I_\text{tr}, I_\text{te}) \cdot p(x\vert I_\text{tr}),
\end{align}
\end{theorem}
where $\lambda_0(x, I_\text{tr}, I_\text{te}) \defeq \frac{p(m_0\vert I_\text{tr})}{p(x\vert I_\text{te})}$ and $\lambda_1(x, y, I_\text{tr}, I_\text{te}) \defeq \frac{p(m_1\vert I_\text{tr})p(y\vert x, m_1, I_\text{tr})}{p(y\vert x, I_\text{te})p(x\vert I_\text{te})}$. This means the weight function $g(x, y, I_\text{tr}, I_\text{te})$ is a reweighing of the original $p(x\vert I_\text{tr})$. The commonly used augmentation can be viewed as a sample-level adjustment to the weight function. From Theorem~\ref{thm_aug_class} we know that the sum of $\lambda_0$ and $\lambda_1$ need not be 1, which is different from some existing augmentation approaches~\citep{zhang2017mixup,yao2022improving}~\footnote{We consider the reformulation of the objectives in~\citet{zhang2017mixup,yao2022improving} according to~\citet{han2022umix}, where the objective with the mixup random variable $\tilde{y}$ can be transformed to the mixup of two weighted terms with the random variable $y$. See Sec. 3.2 of ~\citep{han2022umix} for details.}. The theorem also offers statistical rationale on why the weighted linear combination works~\citep{han2022umix}. Since both $\lambda_0$ and $\lambda_1$ depend on the data statistics from the testing set, methods that utilize sample-independent coefficient~\citep{zhang2017mixup,yao2022improving} should experience degraded performance. We believe this provides insights into the advancement of augmentation-based techniques in the future.

\section{Experiment}\label{sec_exp}

We compare different DBCM variants (see Sec.~\ref{sec_bcm}) benchmarking models with three benchmarking datasets. We showcase the SOTA performance of our models, to demonstrate the consistency of the theory developed in Sec.~\ref{sec_method}. In addition, we provide experimental evidence that complements the theory on explaining why existing works would sacrifice average accuracy for higher worst group accuracy.

\begin{table*}[ht]
\centering
\begin{adjustbox}{width=.9\textwidth,center}
\begin{tabular}{@{}lcccccccc@{}}
\toprule
 & \multicolumn{2}{c}{ColorMNIST(0.5\%)} & \multicolumn{2}{c}{ColorMNIST(2\%)} & \multicolumn{2}{c}{Waterbirds} & \multicolumn{2}{c}{CivilComments} \\
\cmidrule(lr){2-3} \cmidrule(lr){4-5} \cmidrule(lr){6-7} \cmidrule(l){8-9}
& average & worst & average & worst & average & worst & average & worst \\
\midrule
ERM & 81.69 $\pm$ 0.10 & 1.14 $\pm$ 0.40 & 95.23 $\pm$ 0.07 & 56.82 $\pm$ 0.23 &  88.25 $\pm$ 0.16 & 67.76 $\pm$ 0.30 & 87.59 $\pm$ 0.38 & 48.17 $\pm$ 2.61 \\
Mixup~\cite{zhang2017mixup} & 81.12 $\pm$ 2.20 & 0.00 $\pm$ 0.00 & 96.09 $\pm$ 0.20 & 80.00 $\pm$ 2.22 & 88.52 $\pm$ 0.22 & 59.97 $\pm$ 2.01 & 87.67 $\pm$ 0.12 & 53.10 $\pm$ 2.11\\
LISA~\cite{yao2022improving} & 89.45 $\pm$ 1.57 & 21.50 $\pm$ 8.51 & 97.32 $\pm$  0.37 & 87.27 $\pm$ 6.55 & 93.63 $\pm$ 0.66 & 76.95 $\pm$ 4.25 & 87.22 $\pm$ 0.13 & 40.62 $\pm$ 4.32\\
JTT~\cite{liu2021just} & 81.98 $\pm$ 1.17 & 2.00 $\pm$ 0.08 & 95.03 $\pm$ 0.10 & 56.82 $\pm$ 2.21 & 88.32 $\pm$ 0.20 & 68.80 $\pm$ 2.99 & 87.78 $\pm$ 0.29 & 47.06 $\pm$ 2.94 \\
Focal Loss~\cite{lin2017focal} & 67.37 $\pm$ 0.44 & 0.00 $\pm$ 0.00 & 94.62 $\pm$ 0.25 & 43.00 $\pm$ 2.33 & 87.75 $\pm$ 0.36 & 54.67 $\pm$ 2.67 & 87.74 $\pm$ 0.16 & 43.73 $\pm$ 3.66 \\
GroupDRO~\cite{sagawa2019distributionally} & 82.88 $\pm$ 0.09 & 9.00 $\pm$ 0.08 & 95.19 $\pm$ 1.01 & 40.91 $\pm$ 1.20 & 92.03 $\pm$ 0.16 & \textbf{83.64} $\pm$ 1.88 & 86.78 $\pm$ 0.18 & 56.51 $\pm$ 1.93 \\
MMD~\cite{li2018domain} & 11.35 $\pm$ 1.30 & 0.00 $\pm$ 0.00 & 11.35 $\pm$ 2.26 & 0.00 $\pm$ 0.00 & 88.33 $\pm$ 0.51 & 53.58 $\pm$ 2.38 & 82.08 $\pm$ 0.63 & 0.00 $\pm$ 0.00 \\
ReSample~\cite{japkowicz2000class} & 94.95 $\pm$ 0.19 & 66.37 $\pm$ 2.33 & 98.34 $\pm$ 0.23 & 92.00 $\pm$ 1.66 & 93.72 $\pm$ 0.22 & 80.69 $\pm$ 1.86 & 84.59 $\pm$ 1.23 & \textbf{62.17} $\pm$ 1.72 \\
ReWeight~\cite{japkowicz2000class} & 92.43 $\pm$ 0.21 & 57.84 $\pm$ 1.78 & 97.83 $\pm$ 0.19  & 91.46 $\pm$ 1.80 & 93.86 $\pm$ 0.30 & 81.15 $\pm$ 2.20 & 87.04 $\pm$ 0.74 & 58.27 $\pm$ 2.14 \\
\midrule
DBCM(Sec.~\ref{sec_known_s}, known $s$) & \textbf{96.67} $\pm$ 0.27 & \textbf{84.62} $\pm$ 2.02 & \textbf{98.76} $\pm$ 0.20 & \textbf{92.31} $\pm$ 1.73 & \textbf{94.01} $\pm$ 0.19 & 83.18 $\pm$ 2.00 & \textbf{87.85} $\pm$ 0.15 & 43.33 $\pm$ 2.15\\
\bottomrule
\end{tabular}
\end{adjustbox}
\caption{Results on the three benchmarking datasets with accessible attribute $s$. We report both average and worst group accuracy, with mean and standard deviation (``$\pm$") for each of the considered methods after 3 independent runs. The \textbf{boldfaced} values indicate the highest accuracy in comparison.}
\label{tab_known_s}
\end{table*}

\subsection{Experimental Setup}\label{sec_exp_setup}
To ensure a fair comparison, we consider models and datasets prepared by~\citet{yang2023change}. Specifically, we consider two vision datasets: \texttt{Waterbirds}~\citep{sagawa2019distributionally} and \texttt{ColorMNIST}~\citep{nam2020learning,tsirigotis2024group}, and one language dataset: \texttt{CivilComments}~\citep{borkan2019nuanced}, in order to cover the two popular data types. We modify the \texttt{ColorMNIST} dataset such that it aligns with the setup in~\citet{tsirigotis2024group}, which is a harder setup. This is because the vanilla version in~\citet{yang2023change} consists of only two types of attributes, whereas the version in~\citet{nam2020learning,tsirigotis2024group} contains 10 attributes. The modified \texttt{ColorMNIST} contains a ``ratio'' indicator that specifies the portion of samples that do not correlate with labels and attributes. In our experiment, we consider ratios 2\% and 0.5\%, as they are the intermediate and the hardest setups. In practice, we also treat \(p(m_1|I_{\text{tr}})\) and \(p(m_0|I_{\text{tr}})\) serve as prior knowledge/hyperparameters of training composition. Specifically for \texttt{ColorMNIST}, where spurious sample ratio is known, we directly assign 0.5\% or 2\% for \(p(m_0|I_{\text{tr}})\) (i.e., $1 - p(m_1|I_{\text{tr}})$). When the composition ratio is unknown, \(p(m_0|I_{\text{tr}})\) is treated as a hyperparameter and empirically we identify \(p(m_0|I_{\text{tr}}) = 0.85\) performed well across datasets.

\begin{table*}[htp]
\centering
\begin{adjustbox}{width=0.9\textwidth,center}
\begin{tabular}{@{}lcccccccc@{}}
\toprule
 & \multicolumn{2}{c}{ColorMNIST(0.5\%)} & \multicolumn{2}{c}{ColorMNIST(2\%)} & \multicolumn{2}{c}{Waterbirds} & \multicolumn{2}{c}{CivilComments} \\
\cmidrule(lr){2-3} \cmidrule(lr){4-5} \cmidrule(lr){6-7} \cmidrule(l){8-9}
& average & worst & average & worst & average & worst & average & worst  \\
\midrule
ERM & 81.69 $\pm$ 0.10 & 1.14 $\pm$ 0.40 & 95.23 $\pm$ 0.07 & 56.82 $\pm$ 0.23 &  88.25 $\pm$ 0.16 & 67.76 $\pm$ 0.30 & 87.59 $\pm$ 0.38 & 48.17 $\pm$ 2.61 \\
Mixup~\cite{zhang2017mixup} & 81.03 $\pm$ 2.30 & 0.00 $\pm$ 0.00 & 95.26 $\pm$ 0.17 & 42.05 $\pm$ 3.61 & 90.65 $\pm$ 0.30 & 67.29 $\pm$ 1.93 & 87.48 $\pm$ 0.11 & 54.84 $\pm$ 2.13 \\
LISA~\cite{yao2022improving} & 68.09 $\pm$ 2.06 & 0.00 $\pm$ 0.00 & 94.46 $\pm$ 0.53 & 15.91 $\pm$ 13.11 & 89.80 $\pm$ 1.11 & 66.82 $\pm$ 3.87 & 87.18 $\pm$ 0.28 & 49.21 $\pm$ 2.11 \\
JTT~\cite{liu2021just} & 81.80 $\pm$ 0.19 & 2.00 $\pm$ 0.08 & 95.42 $\pm$ 0.02 & 48.86 $\pm$ 1.85 & 88.83 $\pm$ 0.28 & 66.36 $\pm$ 3.10 & 87.78 $\pm$ 3.84 & 47.06 $\pm$ 8.09 \\
Focal Loss~\cite{lin2017focal}  & 67.12 $\pm$ 0.50 & 0.00 $\pm$ 0.00 & 94.53 $\pm$ 0.32 & 37.00 $\pm$ 4.10 & 89.92 $\pm$ 0.43 & 61.68 $\pm$ 3.01 & 87.74 $\pm$ 0.12 & 50.08 $\pm$ 4.10\\
ReSample~\cite{japkowicz2000class} & 81.55 $\pm$ 0.21 & 0.00 $\pm$ 0.00 & 95.70 $\pm$ 0.15 & 65.00 $\pm$ 1.63 & 87.99 $\pm$ 0.12 & 64.17 $\pm$ 1.98 & 83.24 $\pm$ 1.73 & \textbf{68.91} $\pm$ 4.51 \\
ReWeight~\cite{japkowicz2000class} & 76.77 $\pm$ 0.37 & 0.00 $\pm$ 0.00 & 94.93 $\pm$ 0.08 & 54.55 $\pm$ 0.10 & 87.81 $\pm$ 0.18 & 67.60 $\pm$ 1.67 & 87.02 $\pm$1.12 & 58.73 $\pm$ 4.60 \\
\midrule
DBCM(Sec.~\ref{sec_unknown_s_same_data}, $\mathcal{D}_\text{tr} \sim \mathcal{D}_\text{va}$) & \textbf{94.63} $\pm$ 0.35 & \textbf{57.95} $\pm$ 2.30 & \textbf{97.64} $\pm$ 0.10 & \textbf{81.00}  $\pm$ 1.40 & 88.25 $\pm$ 0.05 & \textbf{70.56} $\pm$ 0.12 & \textbf{87.86} $\pm$ 0.30 & 43.41 $\pm$ 2.20 \\
DBCM(Sec.~\ref{sec_unknown_s_different_data}, $\mathcal{D}_\text{tr} \nsim \mathcal{D}_\text{va}$) & 86.12 $\pm$ 0.29 & 3.41 $\pm$ 0.70 & 96.08 $\pm$ 0.20 & 61.36 $\pm$ 1.90 & \textbf{91.04} $\pm$ 0.07 & 62.77 $\pm$ 0.10 & 87.62 $\pm$ 0.20 & 53.89 $\pm$ 2.16 \\
\bottomrule
\end{tabular}
\end{adjustbox}
\caption{Results on the three benchmarking datasets without accessible attribute $s$. We report both average and worst group accuracy, with mean and standard deviation (``$\pm$") for each of the considered methods after 3 independent runs. The \textbf{boldfaced} values indicate the highest accuracy in comparison.}
\label{tab_unknown_s}
\end{table*}

The evaluation consists of 8 benchmarking models from~\citet{yang2023change} that fall into the 4 different classes (see Sec.~\ref{sec_intro} and~\ref{sec_connection}): Mixup~\cite{zhang2017mixup}; LISA~\cite{yao2022improving}; JTT~\cite{liu2021just}; Focal Loss~\cite{lin2017focal}; GroupDRO~\cite{sagawa2019distributionally}; MMD~\cite{li2018domain}; ReSample~\cite{japkowicz2000class}; ReWeight~\cite{japkowicz2000class}.
For each model, we consider two setups, where the first allows the presence of attributes and the second does not. We retrain all the considered models from~\citet{yang2023change} and pick the best models according to the average validation accuracy, which is different from the worst-group-accuracy criterion in~\citet{yang2023change} to match the objective in Eq.~\ref{eq_adjust_obj} (i.e. the framework considers on average accuracy by design). For model optimization, we consider default optimizers and learning rates in~\citet{yang2023change}. Details are deferred to Appendix~\ref{sec_ap_train_config}. Code access: \url{https://github.com/skyve2012/DBA}.

\subsubsection{Attribute $s$ is Known}\label{sec_exp_known_s}
This section presents results with accessible attribute $s$. In addition to the 8 benchmarking models, we also include results with ERM as the baseline. We consider the DBCM variant in Sec.~\ref{sec_known_s}. Results are summarized in Table~\ref{tab_known_s}. It is clear that when the attribute $s$ presents, the proposed DBCM(Sec.~\ref{sec_known_s}, known $s$) achieves the highest average accuracy among all the considered datasets. And all the accuracy of our model exceeds the ERM baseline.  This provides the empirical evidence for Theorem~\ref{thm_adjust_obj_general} and~\ref{thm_adjust_obj_assumption}. Although there is no theoretical quantification on the worst group accuracy, DBCM achieves two highest and one competing (i.e., Waterbirds) worst group accuracy.

\subsubsection{Attribute $s$ is Unknown}\label{sec_exp_unknown_s}
This section presents results without accessing the attribute $s$. We omit results for GroupDRO~\cite{sagawa2019distributionally},  MMD~\cite{li2018domain} as both methods naturally require the knowledge of $s$~\citep{yang2023change}. DBCM(Sec.~\ref{sec_unknown_s_same_data}, $\mathcal{D}_\text{tr} \sim \mathcal{D}_\text{va}$) and DBCM(Sec.~\ref{sec_unknown_s_different_data}, $\mathcal{D}_\text{tr} \nsim \mathcal{D}_\text{va}$) are two variants of the proposed method.
Results are summarized in Table~\ref{tab_unknown_s}. We observe that the proposed DBCM variants achieve the highest average accuracy among all the compared datasets, and 3 out of 4 highest worst group accuracy, suggesting the validity of the methods when $s$ is unknown. And in the case of the worst group accuracy for ColorMNIST(0.5\%), almost all but DBCM cannot correctly classify the worst group samples (i.e. worst group accuracy = 0), suggesting that DBCM method is robust to the change of spurious association between the attributes and the labels.

\subsection{Observation on the Degraded Average Accuracy}\label{sec_shorten_degraded}
From Table~\ref{tab_known_s} and~\ref{tab_unknown_s} we empirically identify an interesting phenomenon---compared to all other methods, DBCM is the only model that consistently outperforms the results of ERM. This observation is aligned with~\citet{yang2023change,tsirigotis2024group}. Yet the previous work did not provide systematic reasoning on why. We argue that the cause is an incorrect model objective that is different from the data composition in $\mathcal{D}_\text{te}$. Specifically, the reduced average accuracy is the result of the misspecified $p(x, y\vert I_\text{te})$ in $g(x, y, I_\text{tr}, I_\text{te})$. For the full explanation, please refer to Appendix~\ref{sec_dis_worse_agv_acc}. It is noteworthy that we are the first to provide such a statistical interpretation of the degradation phenomenon.

\section{Conclusion}\label{seq_conclusion}

In summary, we present the DBA framework to identify the true model objective that improves the test performance. The paper proposes different DBCM variants with weaker assumptions compared to the existing works and demonstrates the SOTA performance. Additionally, we reinterpret the existing work with the proposed framework, which explains the issue of the degraded average accuracy. With the analysis, we convey a message that to achieve decent test performance (even without the access to test during training), one must comprehensively investigate the relationship between those datasets and the model objective. For this purpose, we hope the proposed framework could act as a complementary tool to all the existing work, help people analyze such gaps, and facilitate the development of the corresponding model solutions.

\newpage
\clearpage
\section*{Acknowledgments}
This research was partially supported by Alfred P. Sloan foundation and NSF $\#$1934757.

\bibliography{ref}

\newpage
\clearpage
\appendix

\onecolumn 
\section*{Appendix}

\section{Supplementary Related Work}\label{sec_related_work}
\subsection{Importance Sampling}
DBA interprets the distributional shift as the mismatch of the weight function from the importance sampling perspective. Albeit DBA primarily focuses on the subpopulation setup, we want to point out that the formulation in the method section is general enough to be applied to any of the distributional shift problems.

The discussion on the use of importance sampling for boosting the testing performance can be traced back to 2000 when the authors of the paper discussed the use of importance sampling to improve the accuracy of the test set~\cite{shimodaira2000improving}. The paper focuses on correcting the shifts using the weights as a function of the data distribution and provides theoretical quantification on such shifts relative to the specification of the weights. However, the lack of real-world experiments and the requirement of second-order derivative calculation w.r.t. the model weights limit its application in complex models. Another early endeavor considers a similar importance weighting approach for correcting the error caused by the dataset shifts~\cite{huang2006correcting}. However, the paper only presents a general setup without any assumptions about the data or the model, leaving a wide space of uncleanness in the subpopulation use cases. In comparison, we provide a systematic formulation of how this tool can be applied to the subpopulation problems and explicitly state the underlying assumptions. This offers a clear theoretical ground for identifying the root components (e.g. distributions) that lead to the subpopulation problem and the corresponding solutions. 

Another line of work such as~\citet{kanamori2009least} and~\citet{fang2020rethinking}, provide different ways to estimate the weight function for correcting the shift. Nonetheless, they require partial access to the test set, distinguishing their methods from the proposed one in this paper. 

Besides,~\citet{kimura2024short} offers a comprehensive summary on how important sampling can be applied to solve problems like distributional shift, active learning, model calibration, etc.~\citet{byrd2019effect} studies the importance weighting--a different name but inherently it is just importance sampling--and its impact on the test performance. However, one major limitation of the two papers is that both works consider the weight function only on the data variable $x$, making it insufficient for cases such as subpopulation and covariate shifts, where the attribute and label information are crucial. In comparison, DBA offers a more realistic way with the consideration of the weight function being the ratio of joint distributions of $x$ and $y$. We further provide experimental evidence on the soundness of the proposed framework in Sec.~\ref{sec_exp}.

\subsection{Subpopulation Survey}
Several survey papers provide the summary of subpopulation methods~\footnote{We start with paper~\citep{yang2023change} to identify other existing survey work as this paper is the latest work that conducts comprehensive experimental analysis on existing subpopulation methods. We investigate all the papers it cites and all papers that cite this one to guarantee the coverage and scope as much as possible.}. To the best of our knowledge,~\citet{yang2023change} is the first and possibly the only survey paper that provides a comprehensive experimental study on existing subpopulation methods. The paper decomposes the distribution of $y\vert x$ to take into account the effect of attributes (i.e. spurious features) via Bayes' theorem, where $y$ is the random variable for the labels and $x$ indicates the random variable for data (see Eq. (1) in~\citep{yang2023change} for details). In particular, the paper categorizes 12 different datasets into 4 classes that have different correlations between the labels and the attributes, and benchmarks 20 popular subpopulation methods with these datasets, to characterize the performance of the methods given different attribute setups. Nonetheless, the statistical quantification of why such differences exist is still missing in the paper.

\citet{yu2024survey} and~\citet{2312.16243}, on the other hand, provide a good overview of the existing out-of-distribution (OOD) and domain generalization (DG) methods, which is a superset of the subpopulation methods. While the first paper summarizes the existing works on the application level, the second provides statistical quantification of the error inflation when the distribution shifts are present. One major limitation in the analysis of the second paper is its inability to describe how much inflation can be corrected by model design. It only provides insights on data collection.

Our work, in comparison, can be considered as an extension of the previous work, to offer formal statistical analysis on quantifying the source of error from both modeling and data perspectives. In addition, the proposed DBA can formally answer why the existing methods obtain high worst-case accuracy at the cost of lowering average accuracy on the testing set.

\subsection{Subpopulation Method}

As presented earlier, different subpopulation methods can be categorized into 4 general classes (auxiliary losses, data augmentations, modeling objectives, and data sampling techniques). We summarize these methods in this section for a detailed overview.

In the auxiliary loss class,~\citet{li2018domain} uses maximum discrepancy distance and generative adversarial models to remove the effect of background on the prediction performance. Aiming the same goal,~\citet{arjovsky2019invariant} utilizes a gradient regulation on the last layer of the prediction to enforce consistent perdition across different environments/backgrounds.~\citet{alshammari2022long}, on the other hand, directly adjusts the inference network with a class-balanced distribution. In the data augmentation class, ~\citet{han2022umix},~\citet{zhang2017mixup} and~\citet{yao2022improving} employ convex combination between two arbitrarily drawn data samples, to reduce the effect of the spurious backgrounds. They differ only by the means of generating the coefficient for the convex combination. In the data sampling class, both~\citet{labonte2024towards} and~\citet{izmailov2022feature} look for a subset of class-balanced samples with independent spurious backgrounds. The models are finetuned with this cleaned dataset for optimal performance. The last modeling objective class contains methods that are significantly different from the previous 3 classes. For instance,~\citet{japkowicz2000class} adjust the data by class-balanced weights;~\citet{liu2021just} and~\citet{nam2020learning} train the same model twice, correcting the second model with the incorrect prediction from the first;~\citet{sagawa2019distributionally} proposes the ERM with the focus on the worst-class;~\citet{wu2023discover} proposes a model with two parallel processors, where one discovers the subpopulation and the other corrects it;~\citet{asgari2022masktune} first trains a model, then mask out the learned features from the train model and finetune it the second time, forcing the model to learn robust features during prediction;~\citet{rudner2024mind} utilizes a prior distribution on the subpopulation to improve model performance;~\citet{hong2023harmonizing} and~\citet{han2024improving} implement different loss terms to mitigate the effect of subpopulation on the model. The former uses mutual information whereas the latter considers KL divergence.~\citet{lin2017focal} proposes a focusing parameter to the cross entropy, to reduce the class imbalance impact.

Compared to the existing work, DBA merits its niche. First, it acts as a tool to help people understand the existing methods described above. Specifically, we explicitly state the assumptions on the data, revealing that different existing methods require different underlying assumptions (see Sec.~\ref{sec_connection}). For example, methods that utilize data augmentation~\citep{zhang2017mixup,yao2022improving,han2022umix} assume that training data are conditionally identical given different subpopulation conditionals.~\citet{tsirigotis2024group} and~\citet{menon2020long} requires a uniformity assumption on the label $y$ in the training data. Interestingly, we also discover that all methods can be viewed as having a universal assumption on the identical conditional generative model of data $x$ across all subpopulations. However, none of the previous works states these assumptions explicitly. We believe such discussion is beneficial for people who proceed with this subpopulation direction as for the first time in the community, we explicitly analyze the connections between methods in the same statistical framework.

In practice, DBA outperforms the SOTA benchmarking methods on 3 datasets with 3 proposed methods in the subpopulation setup. This further confirms that DBA is the right framework for this type of problem. More importantly, they are extremely simple to implement due to the nature of importance sampling.

\section{Proofs}\label{ap_proof}

\subsection{Claim~\ref{thm_test_val}}\label{ap_proof_test_val}

We evaluate the chosen model $q^*$ on the testing set $\mathcal{D}_\text{te}$ with the following objective:
\begin{align}\label{eq_proof_test_val}
& \mathbb{E}_{(x, y) \sim p(x, y\vert I_\text{te})} [\log q^*(y\vert x, I_\text{tr})] \nonumber \\
= & \mathbb{E}_{(x, y) \sim p(x, y\vert I_\text{te})} [\frac{p(x, y\vert I_\text{va})}{p(x, y\vert I_\text{va})} \log q^*(y\vert x, I_\text{tr})] \nonumber \\
= & \mathbb{E}_{(x, y) \sim p(x, y\vert I_\text{va})} [\frac{p(x, y\vert I_\text{te})}{p(x, y\vert I_\text{va})} \log q^*(y\vert x, I_\text{tr})].
\end{align}

Under Assumption~\ref{assume_support}, let $z(x, y, I_\text{va}, I_\text{te}) \defeq \frac{p(x, y\vert I_\text{te})}{p(x, y\vert I_\text{va})}$, which is well-defined. This completes the proof.

\subsection{Claim~\ref{thm_adjust_obj_general}}\label{ap_proof_adjust_obj_general}
We first realize that the notation of finding the model ($q\in \mathcal{M}_\text{tr}$) for the best test set performance can be described by the following optimization problem:
\begin{align}\label{eq_obj_opt_adjust_obj_general}
 \max_{q \in \mathcal{M}_\text{tr}} 
 \mathbb{E}_{(x, y) \sim p(x, y\vert I_\text{te})} [\log q(y\vert x, I_\text{tr})].
\end{align}
With the same proof logic as in Sec.~\ref{ap_proof_test_val}, we obtain the following on the objective $\mathbb{E}_{(x, y) \sim p(x, y\vert I_\text{te})} [\log q(y\vert x, I_\text{tr})]$:
\begin{align}
& \mathbb{E}_{(x, y) \sim p(x, y\vert I_\text{te})} [\log q(y\vert x, I_\text{tr})] \label{eq_proof_adjust_obj_general1} \\
= & \mathbb{E}_{(x, y) \sim p(x, y\vert I_\text{te})} [\frac{p(x, y\vert I_\text{r})}{p(x, y\vert I_\text{tr})} \log q(y\vert x, I_\text{tr})] \nonumber \\
= & \mathbb{E}_{(x, y) \sim p(x, y\vert I_\text{tr})} [\frac{p(x, y\vert I_\text{te})}{p(x, y\vert I_\text{tr})} \log q(y\vert x, I_\text{tr})]. \label{eq_proof_adjust_obj_general2}
\end{align}
Under Assumption~\ref{assume_support}, let $g(x, y, I_\text{tr}, I_\text{te}) \defeq \frac{p(x, y\vert I_\text{te})}{p(x, y\vert I_\text{tr})}$, which is well-defined. Since Eq.~\eqref{eq_proof_adjust_obj_general1} and Eq.~\eqref{eq_proof_adjust_obj_general2} are equal, maximizing either one w.r.t. $q \in \mathcal{M}_\text{tr}$ is equivalent to the maximization of the other. This completes the proof.

\subsection{Theorem~\ref{thm_adjust_obj_assumption}}\label{ap_proof_adjust_obj_assumption}
We prove the form of $g(x, y, I_\text{tr}, I_\text{te})$ (Eq.~\eqref{eq_adjust_obj_assumption}) in this section under Assumption~\ref{assume_support},~\ref{assume_data_gen}, and~\ref{assume_s_uniform}. Without loss of generality, we assume the attribute $s$ is categorical and the density for it is $\frac{1}{L}$ (Assumption~\ref{assume_s_uniform}).

We first consider the expansion of $p(x, y\vert I_\text{te})$ and $p(x, y\vert I_\text{tr})$ in $g(x, y, I_\text{tr}, I_\text{te})$.
\begin{align}\label{eq_obj_obj_assumption_1}
p(x, y\vert I_\text{te}) & = \sum_s p(x\vert y, s, I_\text{te})p(y\vert I_\text{te})p(s\vert y, I_\text{te}) \nonumber \\
& \stackrel{\text{Assumption}~\ref{assume_s_uniform}}{=\!=\!=\!=\!=\!=} 
 \sum_s p(x\vert y, s, I_\text{te})p(y\vert I_\text{te})p(s\vert, I_\text{te}) \nonumber \\
 & = 
 \frac{p(y\vert I_\text{te})}{L} \sum_s p(x\vert y, s, I_\text{te}).
\end{align}
\begin{align}\label{eq_obj_obj_assumption_2}
p(x, y\vert I_\text{tr}) & = \sum_s p(x\vert y, s, I_\text{tr})p(y, s\vert I_\text{tr}).
\end{align}
We further expand $p(y, s\vert I_\text{tr})$ to introduce $p(m=m_0)$ and $p(m=m_1)$.
\begin{align}\label{eq_obj_obj_assumption_3}
& p(y, s\vert I_\text{tr}) \nonumber \\  = & p(m_0)p(y\vert, m_0, I_\text{tr})p(s\vert y, m_0, I_\text{tr}) \nonumber \\
+ & p(m_1\vert I_\text{tr})p(y\vert m_1, I_\text{tr})p(s\vert y, m_1, I_\text{tr}) \nonumber \\
  \stackrel{\text{Assumption}~\ref{assume_proposal_1},~\ref{assume_proposal_2}}{=\!=\!=\!=\!=\!=} &  p(m_1\vert I_\text{tr})\frac{p(y\vert I_\text{tr})}{L} + p(m_1\vert I_\text{tr})p(y\vert m_1, I_\text{tr})\mathbf{1}_{\{y=s\}}.
\end{align}
Then 
\resizebox{\columnwidth}{!}{
\begin{minipage}{\columnwidth}
\begin{align}\label{eq_obj_obj_assumption_4}
g(x, y, I_\text{tr}, I_\text{te})^{-1} 
 = & \frac{p(x, y\vert I_\text{tr})}{p(x, y\vert I_\text{te})} \nonumber \\ & \stackrel{\text{Eq.}~\eqref{eq_obj_obj_assumption_2}, \text{Eq.}~\eqref{eq_obj_obj_assumption_3}}{=\!=\!=\!=\!=\!=}  \sum_{s_0} \frac{p(x\vert y, s_0, I_\text{tr})\left[\frac{p(y\vert I_\text{tr})}{L}p(m_0\vert I_\text{tr}) + p(m_1\vert I_\text{tr})p(y\vert m_1, I_\text{tr})\mathbf{1}_{\{y=s\}} \right]}{\frac{p(y\vert I_\text{tr})}{L}\sum_sp(x\vert s, y, I_\text{te})} \nonumber \\
 & =  \frac{p(m_1\vert I_\text{tr})p(y\vert m_1, I_\text{tr})p(x\vert y, s_0=y, I_\text{tr})}{\frac{p(y\vert I_\text{tr})}{L}\sum_sp(x\vert s, y, I_\text{te})} + p(m_0\vert I_\text{tr})\sum_{s_0} \frac{p(s\vert y, s_0, I_\text{tr})}{\sum_s p(s\vert y, s, I_\text{te})} \nonumber \\
& \stackrel{\text{Assumption}~\ref{assume_data_gen}}{=\!=\!=\!=\!=\!=} \frac{p(m_1\vert I_\text{tr})p(y\vert m_1, I_\text{tr})p(x\vert y, s_0=y, I_\text{tr})}{\frac{p(y\vert I_\text{tr})}{L}\sum_sp(x\vert s, y, I_\text{te})} + p(m_0\vert I_\text{tr}) \cdot 1 \nonumber \\
& \stackrel{\text{Assumption}~\ref{assume_data_gen}}{=\!=\!=\!=\!=\!=} p(m_1\vert I_\text{tr}) \cdot \frac{p(y\vert m_1, I_\text{tr})}{\frac{p(y\vert I_\text{tr})}{L} + \frac{p(y\vert I_\text{tr})}{L} \sum_{s \ne s_0, s \ne y, y=s_0}\frac{p(x\vert y, s, I_\text{te})}{p(x\vert y, s_0=y, I_\text{tr})}} + p(m_0\vert I_\text{tr}).
\end{align}
\end{minipage}
}
From Assumption~\ref{assume_data_gen}, we can also obtain the following equality:
\begin{align}\label{eq_obj_obj_assumption_5}
p(x\vert y, s, I_\text{te}) & = p(x\vert y, s, I_\text{tr}) \nonumber \\
 & = \frac{p(y, s\vert x, I_\text{tr})}{p(y, s\vert I_\text{tr})} \cdot p(x \vert I_\text{tr}) \nonumber \\
& = \frac{p(y, s\vert x, I_\text{tr})p(x\vert I_\text{tr})}{p(m_0\vert I_\text{tr})\frac{p(y\vert I_\text{tr})}{L} + p(m_1\vert I_\text{tr})\mathbf{1}_{\{y=s\}} p(y\vert m_1, I_\text{tr})}.
\end{align}
Combining Eq.~\eqref{eq_obj_obj_assumption_4} and Eq.~\eqref{eq_obj_obj_assumption_5}, we have:
\begin{align}\label{eq_obj_obj_assumption_6}
g(x, y, I_\text{tr}, I_\text{te})^{-1} & = p(m_1\vert I_\text{tr}) \cdot \frac{\frac{L}{p(y\vert I_\text{tr})}p(y\vert m_1, I_\text{tr})}{1 + \sum_{s \ne s_0, s \ne y, y=s_0} \frac{\frac{p(y,s\vert x, I_\text{tr})p(x\vert I_\text{tr})}{p(m_0\vert I_\text{tr})\frac{p(y\vert I_\text{tr})}{L} + p(m_1\vert I_\text{tr})\mathbf{1}_{\{y=s\}}p(y\vert m_1, I_\text{tr})}}{\frac{p(y,s=y\vert x, I_\text{tr})p(x\vert I_\text{tr})}{p(m_0\vert I_\text{tr})\frac{p(y\vert I_\text{tr})}{L} + p(m_1\vert I_\text{tr})p(y\vert m_1, I_\text{tr})}}} + p(m_0\vert I_\text{tr}) \nonumber \\
& = p(m_1\vert I_\text{tr}) \cdot \frac{\frac{L}{p(y\vert I_\text{tr})}p(y\vert m_1, I_\text{tr})}{1 + \sum_{s \ne s_0, s \ne y, y=s_0} \frac{p(y, s,\vert x, I_\text{tr}) \cdot \left[ p(m_0\vert I_\text{tr})\frac{p(y\vert I_\text{tr})}{L} + p(m_1\vert I_\text{tr})p(y\vert m_1, I_\text{tr})  \right]}{p(y, s=y,\vert x, I_\text{tr}) \cdot \left[ p(m_0\vert I_\text{tr})\frac{p(y\vert I_\text{tr})}{L}  \right]}}
 + p(m_0\vert I_\text{tr}) \nonumber \\
 & = p(m_1\vert I_\text{tr}) \cdot \frac{\frac{L}{p(y\vert I_\text{tr})}p(y\vert m_1, I_\text{tr})}{1 + \left[ \frac{p(m_0\vert I_\text{tr})\frac{p(y\vert I_\text{tr})}{L} + p(m_1\vert I_\text{tr})p(y\vert m_1, I_\text{tr})}{p(m_0\vert I_\text{tr})\frac{p(y\vert I_\text{tr})}{L}} \right] \cdot \sum_{s \ne s_0, s \ne y, y=s_0}\frac{p(y, s,\vert x, I_\text{tr})}{p(y, s=y,\vert x, I_\text{tr})}}
 + p(m_0\vert I_\text{tr}) \nonumber \\
  & = p(m_1\vert I_\text{tr}) \cdot \frac{\frac{L}{p(y\vert I_\text{tr})}p(y\vert m_1, I_\text{tr})}{1 + \left[ \frac{p(m_0\vert I_\text{tr})\frac{p(y\vert I_\text{tr})}{L} + p(m_1\vert I_\text{tr})p(y\vert m_1, I_\text{tr})}{p(m_0\vert I_\text{tr})\frac{p(y\vert I_\text{tr})}{L}} \right] \cdot \sum_{s \ne s_0, s \ne y, y=s_0}\frac{p(s\vert y, x, I_\text{tr})}{p(s=y,\vert, y, x, I_\text{tr})}}
 + p(m_0\vert I_\text{tr}) \nonumber \\
& = p(m_1\vert I_\text{tr}) \cdot \left( \frac{\frac{L}{p(y\vert I_\text{tr})} \cdot p(y\vert m_1, I_\text{tr})}{ 1 + \left[ \frac{p(m_0\vert I_\text{tr})\cdot p(y\vert I_\text{tr})/L + p(y\vert m_1, I_\text{tr})}{ p(m_0\vert I_\text{tr})\cdot p(y\vert I_\text{tr})/L} \right] \cdot \frac{1 - p(s=y\vert y, x, I_\text{tr})}{p(s=y\vert y, x, I_\text{tr})}} \right) + p(m_0\vert I_\text{tr}),
\end{align}
which is Eq.~\eqref{eq_adjust_obj_assumption} in Theorem~\ref{thm_adjust_obj_assumption}. Because of the following equality under Assumption~\ref{assume_proposal_1} 
\begin{align}\label{eq_obj_obj_assumption_7}
p(y\vert I_\text{tr}) & = p(m_1\vert I_\text{tr})p(y\vert m_1, I_\text{tr}) + p(m_0\vert I_\text{tr})p(y\vert m_0, I_\text{tr}) \nonumber \\
& \stackrel{\text{Assumption}~\ref{assume_proposal_1}}{=\!=\!=\!=\!=\!=} p(m_1\vert I_\text{tr})p(y\vert m_1, I_\text{tr}) + p(m_0\vert I_\text{tr})p(y\vert, I_\text{tr}),
\end{align}
we obtain $p(y\vert m_1, I_\text{tr}) = \frac{p(y\vert I_\text{tr}) - p(m_0\vert I_\text{tr})\cdot p(y\vert I_\text{tr})}{p(m_1\vert I_\text{tr})}$. This completes the proof. 

\subsection{Theorem~\ref{thm_aux_class}}\label{ap_proof_aux_class}
We prove it using some of the previous results from Sec.~\ref{ap_proof_adjust_obj_assumption}. Combine Eq.~\eqref{eq_obj_obj_assumption_4}, Eq.~\eqref{eq_obj_obj_assumption_5}, and Assumption~\ref{assume_aux_y} and~\ref{assume_aux_maj}, we have the following:
\resizebox{\columnwidth}{!}{
\begin{minipage}{\columnwidth}
\begin{align}\label{eq_ap_aux_class}
g(x, y, I_\text{tr}, I_\text{te})^{-1} & = p(m_1\vert I_\text{tr}) \cdot \frac{p(y\vert m_1, I_\text{tr})}{\frac{p(y\vert I_\text{tr})}{L} + \frac{p(y\vert I_\text{tr})}{L} \sum_{s \ne s_0, s \ne y, y=s_0}\frac{p(x\vert y, s, I_\text{te})}{p(x\vert y, s_0=y, I_\text{tr})}} + p(m_0\vert I_\text{tr}) \nonumber \\
& \stackrel{\text{Assumption}~\ref{assume_aux_maj}} {=\!=\!=\!=\!=\!=}  \cdot \frac{p(y\vert m_1, I_\text{tr})}{\frac{p(y\vert I_\text{tr})}{L} + \frac{p(y\vert I_\text{tr})}{L} \sum_{s \ne s_0, s \ne y, y=s_0}\frac{p(x\vert y, s, I_\text{te})}{p(x\vert y, s_0=y, I_\text{tr})}} \nonumber \\
& \stackrel{\text{Eq.}~\ref{eq_obj_obj_assumption_5}, \text{Assumption}~\ref{assume_aux_y}} {=\!=\!=\!=\!=\!=}  \cdot \frac{p(y\vert I_\text{tr})}{\frac{p(y\vert I_\text{tr})}{L} + \frac{p(y\vert I_\text{tr})}{L} \cdot \frac{1 - p(y, s=y\vert x, I_\text{tr})}{p(y, s=y\vert x, I_\text{tr})}} \nonumber \\
& = \frac{1}{\frac{1}{L\cdot p(y, s=y\vert x, I_\text{tr})}} \nonumber \\
& = L\cdot p(y, s=y\vert x, I_\text{tr}).
\end{align}
\end{minipage}
}
Then we have the following from Eq.~\eqref{eq_adjust_obj}:
\begin{align}\label{eq_aux2true_transform}
& \mathbb{E}_{(x, y) \sim p(x, y\vert I_\text{tr})} \left[g(x, y, I_\text{tr}, I_\text{te})\log q(y\vert x, I_\text{tr})\right] \nonumber \\
 = & \mathbb{E}_{(x, y) \sim p(x, y\vert I_\text{tr})} \left[g(x, y, I_\text{tr}, I_\text{te})\log \left( q(y\vert x, I_\text{tr})\cdot \frac{g(x, y, I_\text{tr}, I_\text{te})}{g(x, y, I_\text{tr}, I_\text{te})} \right)\right] \nonumber \\
 = & \mathbb{E}_{(x, y) \sim p(x, y\vert I_\text{tr})} \left[ g(x, y, I_\text{tr}, I_\text{te}) \left( \log q(y\vert x, I_\text{tr}) + \log g(x, y, I_\text{tr}, I_\text{te})^{-1} \right)
 + g(x, y, I_\text{tr}, I_\text{te}) \log g(x, y, I_\text{tr}, I_\text{te}) \right] \nonumber \\
 = & \mathbb{E}_{(x, y) \sim p(x, y\vert I_\text{tr})} \left[ g(x, y, I_\text{tr}, I_\text{te}) \left( \log q(y\vert x, I_\text{tr}) + \log L\cdot p(y, s=y\vert x, I_\text{tr}) \right)
 + g(x, y, I_\text{tr}, I_\text{te}) \log g(x, y, I_\text{tr}, I_\text{te}) \right] \nonumber \\
 = & \mathbb{E}_{(x, y) \sim p(x, y\vert I_\text{tr})} \left[ g(x, y, I_\text{tr}, I_\text{te}) \left( \log q(y\vert x, I_\text{tr}) + \log p(y, s=y\vert x, I_\text{tr}) \right)
 + g(x, y, I_\text{tr}, I_\text{te}) \log L\cdot g(x, y, I_\text{tr}, I_\text{te}) \right].
\end{align}
This completes the proof. 
\subsection{Theorem~\ref{thm_aug_class}}\label{ap_proof_aug_class}

We directly expand the reciprocal of the weight function: 
\begin{align}\label{eq_proof_aug_class_1}
g(x, y, I_\text{tr}, I_\text{te})^{-1} & \defeq \frac{p(x, y\vert I_\text{tr})}{p(x, y\vert I_\text{te})} \nonumber \\
& = \frac{p(m_1\vert I_\text{tr})p(y\vert x, m_1, I_\text{tr})p(x\vert m_1, I_\text{tr}) + p(m_0\vert I_\text{tr})p(y\vert x, m_0, I_\text{tr})p(x\vert m_0, I_\text{tr})}{p(y\vert x, I_\text{te})p(x\vert I_\text{te})} \nonumber \\
& \stackrel{\text{Assumption}~\ref{assume_x_dist_equal}}{=\!=\!=\!=\!=\!=} \frac{p(m_1\vert I_\text{tr})p(y\vert x, m_1, I_\text{tr})p(x\vert I_\text{tr}) + p(m_0\vert I_\text{tr})p(y\vert x, m_0, I_\text{tr})p(x\vert I_\text{tr})}{p(y\vert x, I_\text{te})p(x\vert I_\text{te})} \nonumber \\
& \stackrel{\text{Assumption}~\ref{assume_y_given_x_minor_match}}{=\!=\!=\!=\!=\!=} \frac{p(m_1\vert I_\text{tr})p(y\vert x, m_1, I_\text{tr})}{p(y\vert x, I_\text{te})p(x\vert I_\text{te})} \cdot p(x\vert I_\text{tr}) + \frac{p(m_0\vert I_\text{tr})}{p(x\vert I_\text{te})}\cdot p(x\vert I_\text{tr}).
\end{align}
Let $\lambda_0(x, I_\text{tr}, I_\text{te}) \defeq \frac{p(m_0\vert I_\text{tr})}{p(x\vert I_\text{te})}$ and $\lambda_1(x, y, I_\text{tr}, I_\text{te}) \defeq \frac{p(m_1\vert I_\text{tr})p(y\vert x, m_1, I_\text{tr})}{p(y\vert x, I_\text{te})p(x\vert I_\text{te})}$. This completes the proof.

\section{Restriction and Relaxation of Core Assumptions for Theorem~\ref{thm_adjust_obj_assumption}}\label{sec_relaxation}

This section discusses the limitations of DBCM, aiming to facilitate the research for the future direction. Specifically, we discuss about Assumption~\ref{assume_support} to Assumption~\ref{assume_proposal_2}.

\textbf{Assumption~\ref{assume_support}: }
This assumption requires inclusion relationships on the supports of the training, testing, and validation sets, ensuring a well-defined weight function for importance sampling---a fundamental assumption required for this approach.

\textbf{Assumption~\ref{assume_data_gen}: }this assumption states that the data generation process for \(x\) is identical in both the training and testing datasets, which implies that the underlying generative model for \(x\) given \(y\) remains consistent. This assumption is prevalent in the literature related to disentangled causal processes and conformal inference~\cite{Kang2007DemystifyingDR,tsirigotis2024group,yang2024doubly}. The assumption can be interpreted as a form that characterizes the covariate shift. Relaxing this assumption would require a precise understanding of the relationship between \(p(x|y, s, I_{\text{tr}})\) and \(p(x|y, s, I_{\text{te}})\), which is currently beyond the scope of our work, but could be an interesting avenue for future research.

\textbf{Assumption~\ref{assume_s_uniform}: } This assumption asserts that the label distribution \(x\) does not need to be uniformly distributed, which makes our approach applicable to scenarios with class imbalance. Specifically, this means that the test data can have different class proportions compared to the training data, a common situation in real-world settings. If we treat \(s\) as a treatment variable in causal language, this assumption is akin to requiring strong ignorability between \(x\) and \(s\), with an additional constraint that \(s\) is independent of \(x\). This assumption simplifies the weight function in Theorem 1 by making it depend only on \(p(s=y|y, x, I_{\text{tr}})\). If this assumption were relaxed, we would need additional estimation steps, such as estimating \(p(s=y|y, I_{\text{te}})\).

\textbf{Assumption~\ref{assume_proposal_1}: } This assumption requires that there exists a subset of labels \(y\) that follows the same distribution across both training and testing datasets. Given that we assume inclusive supports (Assumption~\ref{assume_support}) for training and testing datasets, this assumption is not restrictive, as we can subsample the training set to approximate the distribution of labels in the test set.

\textbf{Assumption~\ref{assume_proposal_2}: } This assumption characterizes the nature of the subpopulation shift itself by assuming that the shift is caused by a spurious variable influencing \(Y\). Such a scenario is frequently observed in datasets like Waterbirds, CelebA, and CivilComments~\cite{yang2023change}, where spurious correlations exist between features and labels, leading to subpopulation shifts. This assumption essentially defines the kind of subpopulation shift we are considering, specifically focusing on shifts due to spurious correlations.

\section{Application of DBCM in Difference Scenarios}\label{sec_app_diff_scenarios}

In this section, we outline the three scenarios (see Sec.~\ref{sec_method}) for estimating \(p(s=y|y, x, I_{\text{tr}})\), each applicable based on the availability of certain types of information. Below, we provide practical guidance on selecting the appropriate scenario for a given application:

\textbf{Attribute $s$ is Known: }this scenario applies when \(s\), the spurious variable, is known. Since we have access to \(x\) and \(y\) in the training set, we can directly model \(p(s=y|y, x, I_{\text{tr}})\). This is equivalent to saying that we have perfect knowledge about the source of the subpopulation shift between training and test sets. In such cases, the estimation is straightforward, and the subpopulation shift can be offset using the weight function derived in Theorem~\ref{thm_adjust_obj_assumption}.

\textbf{Attribute $s$ is Unknown, $\mathcal{D}_\text{tr} \sim \mathcal{D}_\text{va}$: }if \(s\) is unknown, but a validation set is available that follows the same distribution as the training set, we use Eq.~\eqref{eq_s_unknow_same_data} for estimating \(p(s=y|y, x, I_{\text{tr}})\). The availability of a validation set allows us to learn about the underlying data distribution and estimate the relevant quantities, even without explicit knowledge of \(s\). Practically speaking, when one suspects a subpopulation shift between train and test sets, a validation set can be used to model the shift and adjust accordingly.

\textbf{Attribute $s$ is Unknown, $\mathcal{D}_\text{tr} \nsim \mathcal{D}_\text{va}$: }when neither \(s\) is known nor is a validation set available, and we assume that the train and validation distributions differ, we suggest estimating \(p(s=y|y, x, I_{\text{tr}})\) like the logit adjustment technique~\cite{menon2020long} but with a different objective function (Eq.~\eqref{eq_s_unknown_diff_data}). Although it is used in a different context, the methodology can be adapted for the estimation here, especially when faced with distributional shifts.

\section{Training Configuration}\label{sec_ap_train_config}
For training, we consider stochastic gradient descent (learning rate = 0.001) for the vision datasets and AdamW~\citep{loshchilov2017decoupled} (learning rate = 0.0001) for the language dataset. All models are trained with a single 16 GB NVIDIA V100 GPU for 3 independent runs\footnote{Code Access: \url{https://github.com/skyve2012/DBA}.}. We follow the convention and report the mean and the standard deviation on both the average accuracy and the worst group accuracy.

\section{Discussion on the Degraded Average Accuracy}\label{sec_dis_worse_agv_acc}

From Table~\ref{tab_known_s} and~\ref{tab_unknown_s} we empirically identify an interesting phenomenon---compared to all other methods, DBCM is the only model that consistently outperforms the results of ERM. This observation is aligned with~\citet{yang2023change,tsirigotis2024group}. However, the previous work did not provide systematic reasoning on why.

With the proposed DBA framework, we argue that this phenomenon is introduced by an incorrect model objective that is different from the data composition in $\mathcal{D}_\text{te}$. Essentially, the DBA framework conveys a single message---we need to optimize for what we want~\footnote{Although in the paper we focus on the subpopulation problem, we want to point out that this message should be generalized beyond this problem set.}. 

We first make the connection between the ERM and the proposed optimization (Eq.~\eqref{eq_obj_general_algo}). During optimization, lowering the ERM aims to improve the models' accuracy in the average sense. This is because ERM is an empirical mean estimator of the loss objective at the population level. We can view the proposed as a generalized ERM for an arbitrary weight function $g(x, y, I_\text{tr}, I_\text{te})$. Then in the case of ERM, the assumption $g(x, y, I_\text{tr}, I_\text{te})=1$ is implicitly made. Different specification on $g(x, y, I_\text{tr}, I_\text{te})$ characterizes different relationships between $\mathcal{D}_\text{tr}$ and $\mathcal{D}_\text{te}$, allowing the correction of the subpopulation shifts therein. Nonetheless, in either case, we do not alter our target on the average accuracy as this is constrained by the empirical risks (regardless of the form of $g(x, y, I_\text{tr}, I_\text{te})$), meaning the worst group accuracy is never a causal result of the optimization. Instead, it is a result of optimization with misspecification on  $\mathcal{D}_\text{te}$. For example, ReSample~\cite{japkowicz2000class} aims to maximize the average accuracy for a group-balanced $\mathcal{D}_\text{te}$. In this case, ReSample implicitly assumes that the $\mathcal{D}_\text{te}$ is group-balanced. However, this need not be the case. In the $\texttt{CivilComments}$ case in $\mathcal{D}_\text{te}$, the groups are still imbalanced, suggesting a misspecification of $g(x, y, I_\text{tr}, I_\text{te})$ that ReSample focuses on than what is the groundtruth $g(x, y, I_\text{tr}, I_\text{te})$ in $\mathcal{D}_\text{te}$. Subsequently, we observe the lowered average accuracy in Table~\ref{tab_known_s}. 

Broadly speaking, from the analysis in Sec.~\ref{sec_connection}, we know that most benchmarking methods essentially propose different forms of class-balance recovery~\footnote{Note that GroupDRO~\citep{sagawa2019distributionally} does not recover the class-balance setup but simply improve the worst group accuracy.}. However, since benchmarking datasets $\mathcal{D}_\text{te}$ do not necessarily share the identical class-balance setup (see Table 2 in~\citep{labonte2024towards}), the objective with which the existing methods optimize may introduce misspecification between the true testing data and the data to which model is optimized. In comparison, since DBCM imposes weaker assumptions, it is reasonable to observe improved performance. It is noteworthy that we are the first to provide such a statistical interpretation of the degradation phenomenon.

\end{document}